\documentclass[journal]{IEEEtran}
\IEEEoverridecommandlockouts
\usepackage{cite}
\usepackage{booktabs}
\usepackage{amsmath,amssymb,amsfonts}
\usepackage{algorithmic}
\usepackage{graphicx}
\usepackage{textcomp}
\usepackage{algorithmic}
\usepackage{array}
\usepackage{tabu}
\usepackage{tabularx}
\usepackage{svg}
\newcommand{\eatme}[1]{ }
\usepackage[inline]{enumitem}
\usepackage{float}
\usepackage{cite}
\usepackage{algorithmic}
\usepackage{array}
\usepackage{tabu}
\usepackage{tabularx}
\usepackage{graphicx}
\usepackage{textcomp}
\usepackage{subcaption,siunitx,booktabs}
\usepackage{multirow}
\usepackage{booktabs} 
\usepackage{ragged2e}
\usepackage{tabularx,tabulary}
\usepackage{adjustbox}
\usepackage{caption}
\renewcommand{\thetable}{\arabic{table}}   
\renewcommand{\thefigure}{\arabic{figure}}
\usepackage{xcolor}

\usepackage{makecell}
\usepackage{fixltx2e}

\def\BibTeX{{\rm B\kern-.05em{\sc i\kern-.025em b}\kern-.08em
    T\kern-.1667em\lower.7ex\hbox{E}\kern-.125emX}}

\begin{document}

\title{Dynamic Graph Attention Networks for Travel Time Distribution Prediction in Urban Arterial Roads\\
}

\makeatletter
\newcommand{\linebreakand}{%
  \end{@IEEEauthorhalign}
  \hfill\mbox{}\par
  \mbox{}\hfill\begin{@IEEEauthorhalign}
}

\author{Nooshin Yousefzadeh, Rahul Sengupta, and Sanjay Ranka\\
\textit{Department of Computer and Information Science and Engineering} \\
\textit{University of Florida, Gainesville, FL, USA} \\
\{nooshinyousefzad, rahulseng,sranka\}@ufl.edu}

\maketitle

\begin{abstract}
Effective congestion management along signalized corridors is essential for improving productivity and reducing costs, with arterial travel time serving as a key performance metric. Traditional approaches, such as Coordinated Signal Timing and Adaptive Traffic Control Systems, often lack scalability and generalizability across diverse urban layouts. We propose Fusion-based Dynamic Graph Neural Networks (FDGNN), a structured framework for simultaneous modeling of travel time distributions in both directions along arterial corridors. FDGNN utilizes attentional graph convolution on dynamic, bidirectional graphs and integrates fusion techniques to capture evolving spatiotemporal traffic dynamics. The framework is trained on extensive hours of simulation data and utilizes GPU computation to ensure scalability. The results demonstrate that our framework can efficiently and accurately model travel time as a normal distribution on arterial roads leveraging a unique dynamic graph representation of corridor traffic states. This representation integrates sequential traffic signal timing plans, local driving behaviors, temporal turning movement counts, and ingress traffic volumes, even when aggregated over intervals as short as a single cycle length. The results demonstrate resilience to effective traffic variations, including cycle lengths, green time percentages, traffic density, and counterfactual routes. Results further confirm its stability under varying conditions at different intersections. This framework supports dynamic signal timing, enhances congestion management, and improves travel time reliability in real-world applications.

\end{abstract}

\begin{IEEEkeywords}
Traffic, Urban Corridor, ATSPM, Dynamic Graphs, Graph Neural Networks.
\end{IEEEkeywords}

\section{Introduction}
As urbanization grows, managing congestion along signalized corridors (arterials) becomes increasingly important for smart city transportation systems. Congestion not only hampers productivity and the economy by causing delays but also impacts societal well-being and the environment due to increased emissions. Evaluating urban traffic corridor performance through travel time and waiting time is essential for optimizing traffic flow. Prolonged delays worsen congestion, underscoring the need for innovative solutions to enhance mobility and sustainability in cities.

Several methods can improve urban traffic efficiency. Coordinated Signal Timing prioritizes main routes by synchronizing green lights with real-time demand, reducing wait times. Real-time data from sensors and cameras helps optimize flow and safety. Actuated Signal Timing \cite{lighthill1955kinematic} adjusts lights based on vehicle presence, ideal for low-capacity roads. Adaptive Traffic Control Systems \cite{gartner2002optimized} dynamically adjust signals in high-capacity areas but face challenges like equipment failures. Intelligent Transportation Systems (ITS) integrate technologies like sensors and cameras to synchronize signals and improve flow, though they are not available everywhere. Analytical solutions \cite{liu2009virtual} use virtual probe trajectories with loop detector data to estimate corridor travel time, but reliable probe data is difficult to obtain due to dynamic traffic patterns. Hence, from a practical standpoint, the solutions that rely on highly available data sources, such as loop detector data and signal state data information to capture relevant traffic control parameters (e.g., signal timing plans), traffic pattern parameters (e.g., turning movement counts and driving behavior) are more feasible.

Arterial travel time serves as a crucial Measure of Effectiveness (MOE) is a Key Performance Indicator (KPI), reflecting delays and congestion over urban transportation networks. This measure captures the impacts of variations in traffic signal coordination and fluctuations in demand, offering a dynamic view of system performance. Continuous assessment and monitoring of travel time are essential to ensure the Level of Service (LOS) provided to commuters meets acceptable standards, maintaining reliability and efficiency. By analyzing the full distribution of travel time, transportation professionals gain a statistically richer and more comprehensive understanding of urban corridor performance. This holistic perspective enables data-driven strategies for more effective traffic management, optimization, and planning, ultimately enhancing mobility and reducing congestion in urban environments.

We propose a Fusion-based Dynamic Graph Neural Network (FDGNN) as a fast, standalone tool for real-time travel time distribution assessment in urban corridors. Extending our prior work by Yousefzadeh et al. (2024), FDGNN moves beyond isolated intersections to corridor-wide applications, enabling comprehensive traffic optimization. The framework leverages attentional graph convolution on dynamic and static representations of urban corridors, using a traffic state matrix to uniquely define graph objects for each scenario. Dynamic graph attributes evolve over time, providing a realistic representation of traffic dynamics. Intermediate fusion captures complex dependencies, while the modular architecture supports hierarchical learning by reusing outputs from static GNNs in dynamic GNNs.

Results show FDGNN can accurately estimate arterial travel time in both directions, even under counterfactual scenarios and varying observational interval windows. Performance was tested across diverse scenarios, including variations in cycle length, green time ratios, and traffic volumes.

This model is designed to be generic in the application to an urban corridor with any number of topologies of intersections, while its performance relies on a limited number of easily accessible traffic factors. These characteristics make FDGNN a scalable and robust solution that can potentially hold promise for real-time urban traffic optimization and adaptive control in real-time. This advanced deep learning-based solution can significantly benefit smart city infrastructures by enabling adaptive, scalable traffic solutions responsive to real-world, corridor-level complexities.

The main contributions of this paper are listed in the following:

\begin{itemize}
\item FDGNN uses interval-based traffic volumes from collector roads feeding into an arterial thoroughfare to infer traffic volumes of intervening roads between intersections, solving a graph completion task via an Attentional Graph Neural Network module.
\item FDGNN constructs a corridor state matrix by concatenating multi-directional interval-based traffic volumes with signal timing parameters, uniquely representing traffic corridors as dynamic graphs with time-evolving, direction-wise relationships between intersections.
\item FDGNN leverages Dynamic Attentional Graph Neural Networks with intermediate feature fusion to learn global graph features. This enables richer, context-aware representations that improve the adaptability of our framework to complex traffic scenarios on long arterial roads.
\item FDGNN employs a sequential learning scheme to enable interdependent and hierarchical representation learning across its modular architecture, enhancing overall performance.
\end{itemize}

This study focuses on a 9-intersection urban corridor in a large U.S. metropolitan area, simulated in SUMO for over 100,000 hours of simulation records under diverse traffic conditions and signal timing plans. Two datasets each containing 50,000 exemplars are generated using either real-world or randomly-generate route files. The proposed model with a total number of 59K parameters (0.23 MB size) is fully parallelized by cost-effective GPU computation. Case studies show that the model delivers accurate, reliable traffic signal coordination, robustness to varying traffic conditions, and interpretable modular assessments, applicable to urban corridors with any number of intersections and configurations. Figure \ref{fig:corridor} illustrates the placement of inflow loop detectors, virtualization of real-world Automated Traffic Signal Performance Measures (ATSPMs) on the base map of the microscopic traffic simulator and other traffic control factors considered in the input of our framework.

The rest of the paper is organized as follows: Section~\ref{prelim} presents a brief overview of the background knowledge and key concepts relevant to this work. Section~\ref{proposedmodels} details the architecture of the proposed digital twin model. Section~\ref{datagen} explains the data generation methodology. Section~\ref{expresults} evaluates the performance of our model through a series of experiments. Section~\ref{related} reviews related work, and finally, Section~\ref{conclusion} concludes the paper with key insights and directions for future research.

\begin{figure*}[htbp]
        \centering
        \captionsetup{justification=raggedright,singlelinecheck=false}
        \includegraphics[scale=0.5]{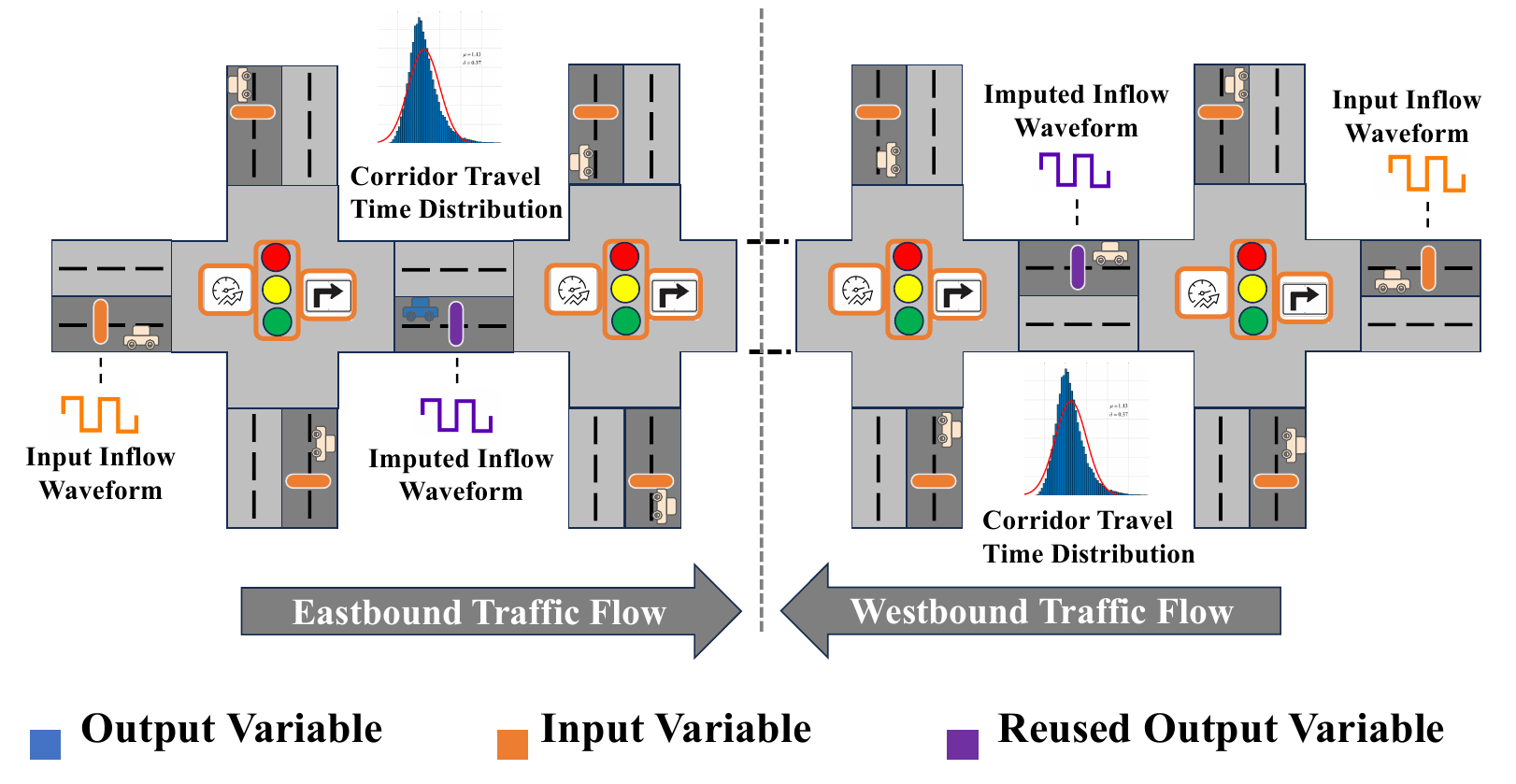}
        \renewcommand{\figurename}{Figure}
        \captionsetup{size=footnotesize}
\footnotesize
        \caption{\textbf{The inputs and outputs used in the modeling of an arbitrary urban corridor.} Inflow loop detectors are positioned 500 meters upstream of intersections. The diagram clearly distinguishes between input variables (in orange), output variables (in blue), and output variables that are reused as inputs in other modules (in purple). Input variables include traffic volume over a certain interval from inflow waveform time series in three directions upstream of an intersection, signal timing parameters (e.g., cycle length, offset, maximum green duration), driving behavior parameters (e.g., speed, acceleration, space cushion, lane changing, etc.), turning movement counts, and the distance between consecutive inflow loop detector locations along the main route. Reused output variables include traffic volume of intervening road segments between intersections on the arterial thoroughfare. Output variables include the normal distribution of eastbound and westbound arterial travel times.}
        \label{fig:corridor}
    \end{figure*}

\section{Preliminaries}
\label{prelim}

\subsection{Graph Neural Networks} 
Graph Neural Networks (GNNs) leverage the inherent structural information of graphs, where nodes represent entities and edges represent relationships. These networks have proven effective in various applications, such as social network analysis, recommendation systems, bioinformatics, and knowledge graph completion. GNNs utilize a message-passing paradigm, where nodes iteratively exchange information with their neighbors to update their representations. This iterative process allows the model to capture hierarchical and recursive patterns in graph-structured data.
A key component of GNNs is the message-passing mechanism, which enables nodes to update their embeddings by aggregating information from their neighbors. This process is mathematically represented by an equation involving an activation function, a weight matrix for each neighborhood, and a bias vector for each node's representation. After several layers of message passing, nodes accumulate information from progressively distant neighbors, enhancing their ability to understand the graph's structure. More specifically, The node representation after the $k^{th}$ layer in a GCN is given by the following formula:

\[
h_i^k = \sigma \left( \sum_{j \in \mathcal{N}(i)} \frac{1}{c_{ij}} W_k h_j^{k-1} \right)
\]

where $c_{ij}$ is a normalization factor defined by the inverse square root of the degrees of nodes $i$ and $j$. Through multiple iterations, GCNs refine node representations by considering information from progressively distant neighbors, up to $k$-hops.

\subsection{Graph Convolutional Networks} 
Graph Convolutional Networks (GCNs) are a specific variant of GNNs, designed to adapt Convolutional Neural Networks (CNNs) to graph data. GCNs aggregate information from neighboring nodes using a weighted sum, where the weights are influenced by the graph structure. The key steps in a GCN include initialization, message aggregation, normalization, pooling, and iteration. Normalization ensures that the gradients don't vanish, and pooling combines the normalized messages with node representations. Through multiple iterations, GCNs refine node representations by considering neighbors up to a specified number of hops. More specifically, The node representation after the $k^{th}$ layer in a GCN is given by the following formula:

\[
h_i^k = \sigma \left( \sum_{j \in \mathcal{N}(i)} \frac{1}{c_{ij}} W_k h_j^{k-1} \right)
\]

where $c_{ij}$ is a normalization factor defined by the inverse square root of the degrees of nodes $i$ and $j$. Through multiple iterations, GCNs refine node representations by considering information from progressively distant neighbors, up to $k$-hops.

\subsection{Dynamic Graph Convolutional Networks}
Dynamic Graph Convolutional Networks (DGCNs) extend traditional GCNs by incorporating dynamic graph structures, where the graph topology or edge features can change over time or in response to specific inputs. This adaptability enables DGCNs to handle evolving relationships between nodes, which is especially useful in scenarios like traffic forecasting, social networks, and recommendation systems. In DGCNs, the graph convolutional operation can be adjusted dynamically at each layer based on the current graph structure, with edge features or node interactions influencing the aggregation process. The update rule for dynamic graph convolutions is formulated as:

\[
h_i^k = \sigma \left( \sum_{j \in \mathcal{N}(i)} \frac{e_{ij}^k}{c_{ij}} W_k h_j^{k-1} \right)
\]

where $e_{ij}^k$ represents dynamic edge features at the $k^{th}$ layer, and $c_{ij}$ is the normalization factor. This formulation allows the model to adapt to changes in the graph, enabling more accurate and flexible learning from dynamic data.

\section{Proposed Models}
\label{proposedmodels}
In this study, we estimate the bidirectional arterial travel time distribution using an approximated and deterministic approach grounded in the normality assumption. This method not only accounts for uncertainty but also enhances the generalization of the original problem. The result is a robust, efficient, and flexible framework for modeling travel time dynamics in urban corridors, facilitating integration with other frameworks, and solving various downstream tasks.

Dynamic Graph Neural Networks (DGNNs) extend conventional Graph Neural Networks (GNNs) by incorporating the ability to process time-evolving data. These networks have demonstrated efficacy in addressing complex real-world problems by leveraging their adaptability, scalability, and the integration of both temporal and spatial information in graph structures \cite{xu2020inductive, seo2018structured}. The intermediate fusion of node and edge representations in graphs enables the extraction of intricate patterns that emerge from evolving structural and feature dynamics, offering a comprehensive understanding of the underlying graph properties.

In this study, we leverage the combination of static and dynamic graphs to estimate the distribution of time it takes to travel through any arterial traffic corridor. This approach leverages the strengths of both types of GNNs in order to capture spatial and temporal complexities in traffic patterns effectively. To enhance representation learning, we introduce an intermediate fusion technique for node and feature embeddings. This fusion captures the interplay between structural configurations and feature evolution, resulting in an enriched representation of the target concepts \cite{yousefzadeh2023comprehensive}.

Furthermore, we adopt a sequential optimization strategy that provides a structured framework for managing the complexities of representation learning through progressive, step-wise decision-making. This approach enables efficient handling of dynamic variations in the underlying data. The proposed Fusion-based Dynamic Graph Neural Networks (FDGNN) framework evaluates urban traffic corridor performance by accurately estimating the distribution of arterial road travel times, derived from input density histograms with a fine resolution of 10-second bins.

The proposed model and input graph data are designed to be generic, making them applicable to any urban traffic corridor regardless of the number or topology of the encompassing intersections.

An overview of the FDGNN framework is depicted in Figure \ref{fig:overview}. Table \ref{table:notation} comprehensively describes the notations, terminologies, and variables used in this paper. The complete source code is publicly available to support further research and application.\footnote{Source code: https://github.com/NSH2022/Traffic-Corridor-NormalPDF}

\begin{table}[ht]
    \begin{adjustbox}{width=\columnwidth,center}
        \begin{tabulary}{1.0\textwidth}{|l|l|l|l|l|l|}
            \hline
            \textbf{Notation}&\textbf{Description}  &\textbf{Aggr}  &\textbf{Size}  &\textbf{Type}  \\ 
            \hline
            $k$& number of intersections&-&1x1&Integer, 0-9\\\hline
            $w$& Size of time series window &-  &1x1  &Integer, 0-360 \\\hline
            $inf$& Count of vehicles upstream the intersection within $w$  &5 sec  &1x$w$  &Integer, 0-8 \\\hline
            $tt$& Probability density values from the normal travel time distribution &10 sec&1xm&Integer, 0-250\\\hline
            $drv$& Driving behavior parameters  &-  &1x9 &Float, 0-30\\\hline
            $tmc$& Turning-movement counts ratio &-  &1x12  &Float, 0-1 \\\hline
            $dis$& Distance to the next intersection &1 m  &1x1  &Float, 0-600 \\\hline
            $cyc$& Common cycle length throughout the corridor &1 sec &1x1&Integer, 150-240 \\\hline
            $maxDr$& Effective green time for the phase &1 sec&8x1&Integer, 0-240\\\hline
            $minDr$& Effective green time for the phase &1 sec&8x1&Integer, 0-240\\\hline
            $off$& Offset in the start of the cycles &1 sec&11&Integer, 0-240\\\hline
            $M_{x}$& GNNs for $inf$ imputation between intersections &-&-&torch.nn.Module\\\hline
            $M_{\mu}$& GNNs module for estimation of mean of $tt$(s)&-&-&torch.nn.Module\\\hline
            $M_{\sigma}$& GNNs module for estimation of standard deviation of $tt$(s) &-&-&torch.nn.Module\\\hline     
        \end{tabulary}
    \end{adjustbox}
\caption{Summary of the notations and their definitions.}
\label{table:notation} 
\end{table}

\subsection{Intermediate Traffic Volumes between Intersections}

The module ($M_{x}$) is a graph neural networks model trained on static graph data for a node feature imputation task. It is a specific form of $M_{ext}$ introduced in our previous work \cite{yousefzadeh2024graph}, which imputes node features as intervening traffic volume between intersections. 

In its general form, this module includes an additional residual attention sub-module for encoding the sequence of input time series as node features for an urban traffic intersection. These features are replaced with traffic volume as scalar vehicle count values at the urban corridor scale in this current work. Hence, in this study, the $M_{x}$ module consists of two Graph Attention Layers (GATs) with 4 and 1 attention heads, respectively. These layers embed 8-dimensional node features into a 32-dimensional hidden representation, followed by a fully connected layer, with all layers activated by the ReLU activation function. This model takes input features in the form of bidirectional acyclic graph objects, as described in Subsection \ref{graphrep}. The function of the module can be defined as follows:
\[
M_{x}: G(A, X_i, E_i) \rightarrow \text{inf}_i^{ej}, \text{inf}_i^{wj}
\]
where $A$ represents the common topology of an urban intersection composed of $K$ consecutive intersections, as described in Subsection \ref{graphrep}.

\subsection{Mean and Standard Deviation of Travel Time Distribution}

The module ($M_{\mu}$ and $M_{\sigma}$) is a graph neural networks model trained on dynamic graph data. Compared to static graphs, dynamic graphs are more informative node features and time-evolving edge features. In this model, we employ an intermediate fusion technique to merge the node and edge feature embeddings into a general graph representation, addressing a complex graph regression problem to estimate the mean ($\mu$) and standard deviation ($\sigma$) of travel time distributions through arterial roads in urban corridors. The function of these two modules can be defined as 
\[
M_{\mu}: G(A, X'_i, E'_i) \rightarrow \mu_{tt_i^{e}}, \mu_{tt_i^{w}}
\]
\[
M_{\sigma}: G(A, X'_i, E'_i) \rightarrow \sigma_{tt_i^{e}}, \sigma_{tt_i^{w}}
\]

The architectures of both $M_{\mu}$ and $M_{\sigma}$ are identical. Each comprises two GAT layers with 4 and 1 attention heads, respectively, embedding 14-dimensional node features into a 64-dimensional hidden representation. This is followed by an edge MLP submodule and two fully connected layers, generating outputs for both forward and reverse directions of movement. All layers use the ReLU activation function. 

The MLP submodule models the time-evolving edge features expected in the dynamic graph representation of traffic in urban corridors. After the first GAT layer, the edge MLP submodule is applied, consisting of two fully connected layers that encode and decode the 19-dimensional edge features into and from a 64-dimensional hidden space, with a ReLU activation in between. The second GAT layer processes the updated edge features output by the MLP submodule along with the node features output by the first GAT layer, maintaining the same edge connectivity. 

As described in Subsection \ref{graphrep}, edge features are direction-specific, while node features are global to the direction of movement. Therefore, the final edge embeddings are first split based on their forward and reverse directions, averaged, and then concatenated with the final global node embeddings. These concatenated features are subsequently fed into two final fully connected layers, activated by ReLU, to produce the model's outputs. 

The modular architecture enhances the network's representation learning, interpretability, and robustness of predictions. Notably, it allows for sequential optimization, which is crucial for this architecture since the outputs of the dynamic GNNs rely on the outputs of the static GNNs.

\subsection{Sequential Optimization Technique}
Many real-world processes inherently follow a normal distribution or a close approximation thereof. The assumption of normality provides a realistic means of generalizing these processes, enhancing the robustness and accuracy of predictive tasks. In this study, the limited number of vehicles in traffic simulations leads to histograms that are often noisy and discontinuous due to sparse data or irregular binning. Approximating such histograms with a discrete normal probability density function (PDF) effectively smooths the distribution, mitigating noise and enabling better generalization in our model.

Furthermore, compared to sparse and noisy histograms, fitted normal PDFs exhibit significantly lower representational complexity. This reduction simplifies the model's parameterization, resulting in improved memory and computational efficiency. Additionally, the simplified framework enhances flexibility, making it more adaptable for downstream tasks while maintaining the integrity of the underlying data representation.

FDGNN framework, encompass previously introduced components $M_{inf}$, $M_{\mu}$ and $M_{\sigma}$ . It takes two  static and dynamic versions of graph representations of the target traffic corridor (see Subsection Graph Data Construction \ref{graphrep}):

\[
FDGNN: G, G' \rightarrow {tt_i^{e}}, {tt_i^{w}}
\]

We implement our framework using PyTorch machine learning framework and PyTorch Geometric library. We employ a sequential optimization strategy within the training loop to optimize our framework. Sequential optimization involves updating the parameters of individual modules one at a time, rather than optimizing the entire network simultaneously. This approach is particularly advantageous when modules exhibit interdependencies or serve distinct roles in the overall network optimization. During each training epoch, multiple optimizer objects are utilized to perform the backward pass and parameter updates for one module at a time. This technique allows each module to be fine-tuned independently, potentially leading to improved convergence and enhanced overall model performance.

The training process begins by extracting the mean and standard deviation of the normal probability density function (PDF) for travel times in the east and west directions from the original travel time histograms. Three independent Adam optimizers are then defined, each with its own learning rate and a mean squared error (MSE) criterion (squared L2 norm) to drive the sequential optimization process.

\noindent\textbf{Step 1:} The first optimizer, $optimizer_{inf}$, is employed to train the static graph model $M_{inf}$ on static graph data with masked node features. The objective is to impute intermediate traffic volumes between intersections during arterial phases.

\noindent\textbf{Step 2:} The imputed traffic volumes from the previous step are then used to replace the node and edge features in the dynamic graph.

\noindent\textbf{Step 3:} The second optimizer, $optimizer_{mean}$, is used to train the dynamic graph model $M_{\mu}$ on a regression task. The goal is to predict the mean values of the bidirectional travel time distribution, $mean_{east}$ and $mean_{west}$.

\noindent\textbf{Step 4:} Finally, the third optimizer, $optimizer_{stdv}$, trains the dynamic graph model $M_{\sigma}$ on a regression task to predict the standard deviation of the bidirectional travel time distribution, $stdv_{east}$ and $stdv_{west}$.
The overall loss function comprises three components, each corresponding to one of the aforementioned tasks and optimized independently. These loss components can be formally defined as follows:

\[
\begin{split}
& \quad loss_{inf} = MSE(\hat{inf}, inf) \\
&\quad loss_{mean} = MSE(\hat{\mu}_{east}, {\mu}_{east}) + MSE(\hat{\mu}_{west}, {\mu}_{west}) \\
&\quad loss_{stdv} = MSE(\hat{\sigma}_{east}, {\sigma}_{east}) + MSE(\hat{\sigma}_{west}, {\sigma}_{west}) \\
\end{split}
\]

This sequential optimization framework ensures that the individual components of the model are effectively trained, leading to a more robust and accurate representation of the travel time dynamics in urban corridors.

\begin{figure*}[!tbp]
  \centering
  \begin{minipage}[b]{0.4\textwidth}
    \includegraphics[width=\textwidth]{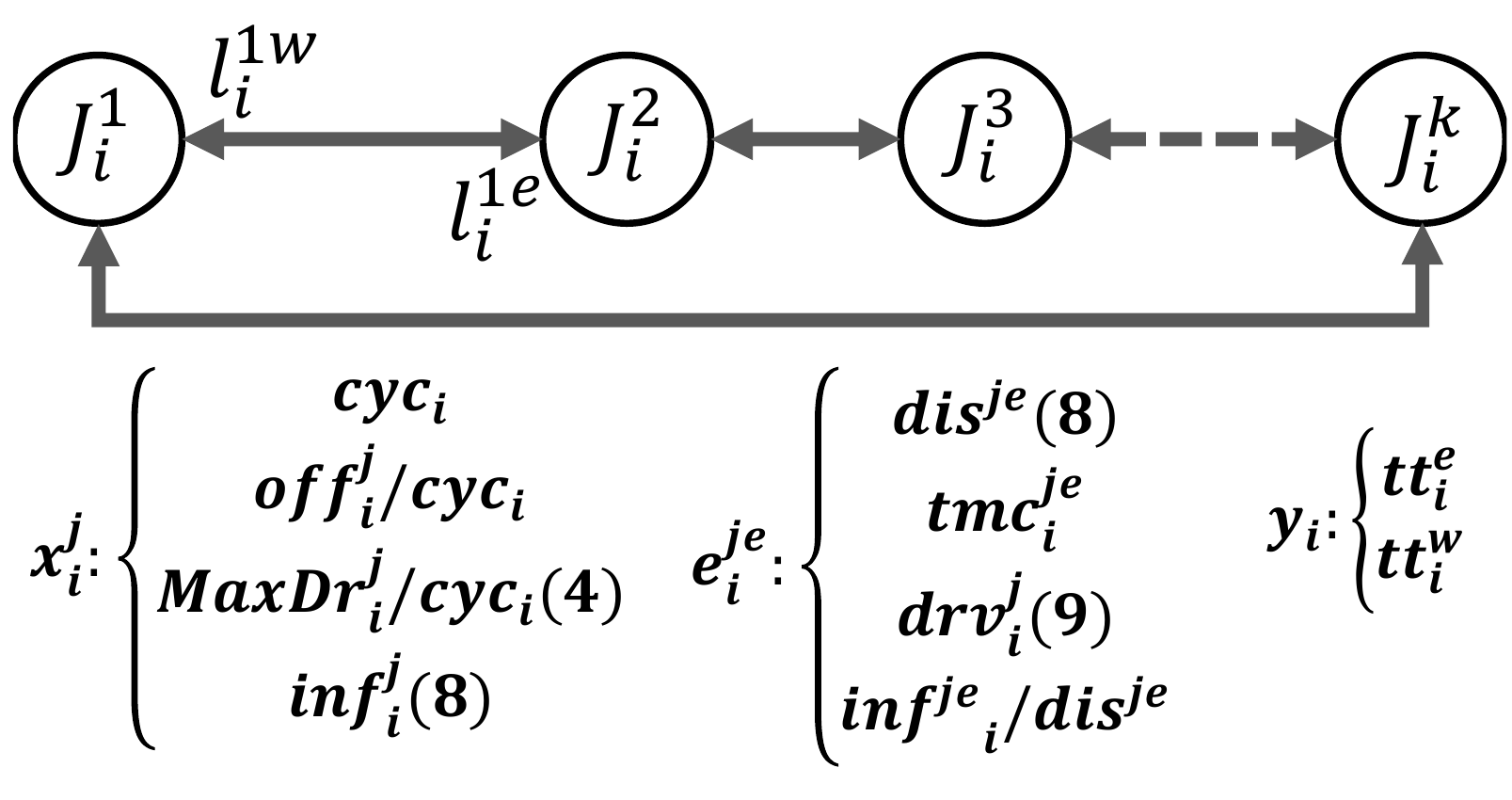}
    \renewcommand{\figurename}{Figure}
    \captionsetup{size=footnotesize}
    \caption{\textbf{Graphical representation of traffic state of an urban corridor.} Important factors are attributed as features to nodes and edges to uniquely represent the traffic state of an arbitrary urban corridor as a bidirectional and acyclic (dynamic) graph with k nodes indexed by $\{J^1,..J^k\}$ and 2k edges. The edge features are attributed to the direction of the movement between intersections. }
    \label{fig:graph}
  \end{minipage}
  \hfill
  \begin{minipage}[b]{0.5\textwidth}
    \includegraphics[width=\textwidth]{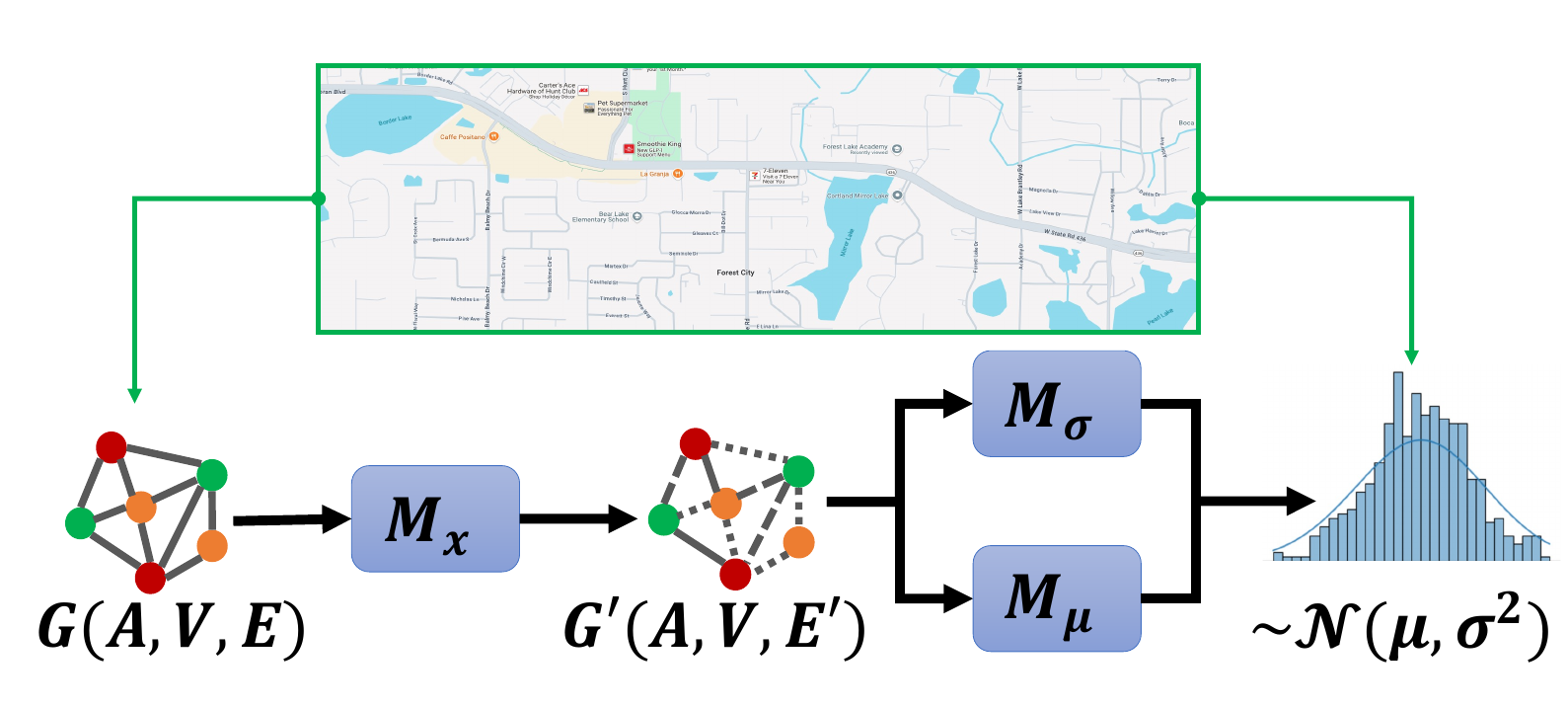}
    \renewcommand{\figurename}{Figure}
    \captionsetup{size=footnotesize}
    \caption{\textbf{Overview of the proposed framework.} This diagram illustrates the architecture of the proposed FDGNN framework, which consists of three modules. The $M_x$ module processes static graph data with masked node features representing intervening traffic volumes, reconstructing the node features. These reconstructed features are then passed as inputs to the $M_{\mu}$ and $M_{\sigma}$ fusion-based modules. The framework's outputs are a discrete normal probability density function (PDF) of bidirectional arterial travel time. The modules are sequentially optimized and the final loss is computed with respect to the fitted travel time histograms obtained from traffic simulations on the target urban corridor.}
    \label{fig:overview}
  \end{minipage}
\end{figure*}

\section{Traffic Data Preparation}
\label{datagen}
In this section, we provide a detailed explanation of how the datasets used in this study were generated and preprocessed. To ensure clarity and simplicity, we divide the data preparation procedure into three key steps: Traffic Simulation, Log Extraction, and Graph Data Construction. Each step is described as follows:

\subsection{Traffic Simulation}
This study is based on realistic traffic data generated by the SUMO (Simulation of Urban MObility) micro-simulator framework, based on real-world parameters including road layouts, signal timings, vehicle flow rates, and traffic rules. The target arterial road contributed in this study features a series of earlier 8 east-west directional intersections on State Road 436 (SR 436) arterial road running from US 441 in Apopka in north of Orlando City. The intersections have unique characteristics e.g., typologies, geometries, lane configurations, traffic rules, traffic patterns, and signal timing plan settings. The extensive dataset includes over 100,000 hours of traffic data simulation using ThreadPoolExecutor for multi-threading in Python while running this large amount of simulations in parallel. We use real-world ATSPM (loop detector) data and sparse WEJO (GPS) probe data to generate an approximate Origin-Destination (OD) probability matrix, providing insights into the likelihood of vehicles traveling between locations. The SUMO tool, od2trips, is employed to create route files for the simulation. The dataset used in this study is \textit{Real-TMC dataset} based on the realistic turning movement counts matrix inferred by approximate OD matrices. However, for comparison purposes, we also generate \textit{Real-TMC dataset} based on randomized route files. Based on the configuration parameters outlined in Table \ref{table:datagen} a python multiprocessing script is used to run a large number of nearly real traffic scenarios on urban traffic corridors under study to generate the required raw data used in this study. 

\subsection{Log Extraction}
A simulation Record is a record of a single run of traffic simulation that is extracted and processed into structured XML log files. These raw log files serve as the intermediary dataset for subsequent analysis to produce a zip file for each traffic simulation record with information tailored to the specific needs of our study. The log extraction process includes several steps including recording of the events, calculation of metrics, data clearing, and data storage. During the metric calculation step, various Python modules are added to operate additional tasks e.g., count the number of vehicles crossing each road segment, computer the traveling time of each vehicle, and generate density histograms for traveling times through certain paths.

Specifically, we use the Floating Car Data (FCD) output log file generated by SUMO to compute travel time at arterial directions of movement as the target variable assigned to each traffic scenario simulation. This log contains the trajectory information for each vehicle during the simulation. By analyzing the vehicle coordinates, we can determine the time it takes for a vehicle to cross a specific region. This is done by filtering the coordinates and their corresponding timestamps to find the vehicle's entry and exit times. We then calculate the corridor travel time by recording the entry time at the first intersection and the exit time at the final intersection. This process is applied separately for both the East-West and West-East routes. Vehicles that do not exit the final intersection are excluded from the corridor travel time calculations.

\subsection{Graph Data Construction}
\label{graphrep}
To enable advanced analysis and data mining with graph neural graph networks, the traffic data has to be transformed into graphs. Further, to improve the expressiveness of data for a comprehensive and accurate assessment of arterial bidirectional travel time, two types of graph (i.e., static, and dynamic) representations are constructed and leveraged throughout the framework pipeline. Both types of graph data objects are bidirectional and acyclic and share a uniform structure comprising 8 nodes corresponding to the eight intersections and 16 edges representing road segments connecting each pair of consecutive intersections in both directions of movement through the arterial road. The use of acyclic graphs is especially useful to avoid inefficiencies or errors, such as infinite loops during traversal. Each node and edge is enriched with features derived from the traffic logs. To better capture the dynamic traffic patterns, we also include time-dependent features to capture changes over time in the form of a temporal graph structure. Node features are directional agnostic, while edge features are directional specific and are attributed specifically to the movement (linkage) direction. 

The static graph representation of the corridor, denoted as \( G(A, X, E) \), serves as input to the module \( M_x \). Here, \( X \in \mathbb{R}^{8 \times 8} \) represents the node features, and \( E \in \mathbb{R}^{16 \times 19} \) denotes the edge features. Corresponding to the $i$\-th traffic scenario simulation sample and the direction of movement $o$, the node feature matrix $X_i = \{(inf_i)\}$  encapsulates the traffic volume associated with each intervening road segment of the eight-intersection corridor, while the edge features $E_i = \{(dis^o, tmc_i, drv_i, inf_i/dis^o)\}$ comprise the following attributes respectively, with respect to the direction of movement:
\begin{itemize}
    \item The fixed distances between consecutive intersections computed at each direction of movement,
    \item Turning movement counts at each intersection,
    \item Driving behavior parameters, and
    \item Traffic density of the road segment corresponding to the direction of movement, calculated as the ratio of traffic volume to the length of the road segment.
\end{itemize}

In this framework, the node feature matrix \( X \) is masked to include only the traffic volume of effective phases along the east-west arterial road. The masking operation is performed through an element-wise multiplication between the node feature matrix and a binary mask matrix \( T \in \{0, 1\}^{8 \times 8} \). This binary matrix ensures that only non-arterial phases of traffic volume are considered during computation.

\[
T_{i,j} =
\begin{cases}
1, & \text{otherwise,} \\
0, & \text{for phase 1,2,5,6.}
\end{cases}
\]

The corridor dynamic graph data \( G'(A, X', E') \) is used as input to the \( M_{\mu} \) and \( M_{\sigma} \) modules and is illustrated in Figure \ref{fig:graph}. Unlike the static graph, the dynamic graph object is constructed to abstract the concept of the traffic corridor holistically. The edge features of the dynamic graph \( E' \in \mathbb{R}^{16 \times 19} \) of the dynamic graph are identical to those in the static graph; however, they are not fixed and evolve over the time. This temporal evolution is modeled within the architecture of the graph neural networks. The node features of the dynamic graph \( X' \in \mathbb{R}^{K \times 14} \), on the other hand, are designed to combine the fixed signal timing state of the corridor with the random inflow traffic pattern state, forming a corridor state matrix that uniquely and momentarily represents the traffic flow in an urban corridor. In general, the corridor state matrix assigned to graph nodes, \( X' \in \mathbb{R}^{8 \times (2 + m/2 + m)} \) aggregate feature vectors from \( K = 8 \) intersections operating in \( m = 8 \) distinct traffic signal phases. More specifically, corresponding to the $i$\-th traffic scenario simulation sample, the node features matrix $X'_i = \{(cyc_i, off_i/cyc_i, MaxDr_i/cyc_i, inf_i)\}$ respectively comprise the following attributes for every intersections:

Each feature vector in the corridor state matrix concatenates:
\begin{itemize}
    \item Traffic signal cycle length,
    \item Offset ratio computed as the ratio of the offset to the length of the cycle,
    \item Effective green light ratio for phases 1, 2, 5, and 6 along the arterial road as a ratio of maximum time of green to the cycle length, and
    \item Traffic volume associated with every traffic signal phase is calculated as the count value of vehicles within a time series window.
\end{itemize}

\[
\text{X'} =
\scalebox{0.8}{$
\begin{bmatrix}
\text{cycle}_1, \text{offset}_1, \text{green}_1^1, \text{green}_1^2, \text{green}_1^5,, \text{green}_1^6, \text{inf}_1^1, \text{inf}_1^2, \ldots, \text{inf}_1^m \\
\text{cycle}_2, \text{offset}_2, \text{green}_2^1, \text{green}_2^2, \text{green}_1^5, \text{green}_2^6, \text{inf}_2^1, \text{inf}_2^2, \ldots, \text{inf}_2^m \\
\vdots \\
\text{cycle}_k, \text{offset}_k, \text{green}_k^1, \text{green}_k^2, \text{green}_1^5, \text{green}_k^6, \text{inf}_k^1, \text{inf}_k^2, \ldots, \text{inf}_k^m
\end{bmatrix}
$}
\]

Regarding the target variable of prediction, first, we convert the resulting density histograms of traveling time(S) to Probability Density Functions (PDFs). Our observation on negligible skewness and kurtosis from a Pearson distribution fit to the original travel time histograms served as evidence to support the assumption that the distribution of travel time in simulated traffic data approximates a normal distribution. This statistical technique is especially useful in eliminating irrelevant variability of bin-specific artifacts of histograms (e.g., variations due to bin size or boundaries), and focusing on the core distribution instead for more robust estimation. Corresponding to the $i$\-th traffic scenario simulation sample, the target PDF(s) of corridor travel time $y_i = \{(tt_i^e, tt_i^w)\}$ can be computed for any desired resolution (i.e., bin size) of output travel time histograms, and the input features can be collected for any desired size of the window over input inflow waveforms time series. The final graph data is serialized as objects of the torch\_geometric.data.Data class and stored in .pth files.

\begin{table*}[!htbp]
\centering
    \begin{adjustbox}{width=\textwidth,center}
    \begin{tabulary}{1.2\textwidth} {|l | L | L|}
    \hline
        \textbf{Parameter} &\hfil \textbf{ Description} &\hfil \textbf{ Variation} \vspace{2mm}\\
        \hline
        \multicolumn{3}{|l|}{\textbf{Signal Timing Plan Parameters}}\\
        \hline
        Total cycle Lengths &
          Length of the common cycle for the corridor &
          Varies from 150 seconds to 240 seconds\\
        Offsets &
          Offsets in the start time of the cycles at various intersections &
          Offsets are varied randomly from 0 to the cycle length of the respective intersections\\
        Barrier Times &
          When the barrier in Ring-and-Barrier occurs, separating the non-coordinated phases and the coordinated phases &
          Occurs randomly while ensuring minimum allowable green times for the phases along with yellow and red times are met.\\
        Phase Duration &
          Length of Green, Yellow, and Red times for the phases making up the dual rings of Ring-and-Barrier operation &
          Minimum and Maximum Green times, Yellow and Red times are fixed based on field settings\\
        Phase Order &
          Phase order in the dual rings of Ring-and-Barrier operation &
          Phase orders fixed for each intersection based on field settings\\
        \hline
        \multicolumn{3}{|l|}{\textbf{Driving Behavior Parameters}}\\
        \hline
        accel &
          SUMO parameter for vehicle acceleration &
          From 1.6 to 3.6 meters per second squared\\
        decel &
          SUMO parameter for vehicle deceleration &
          From 3.0 to 6.0 meters per second squared\\
        emergencyDecel &
          SUMO parameter for maximum possible deceleration for a vehicle &
          From 6.0 to 12.0 meters per second squared\\
        minGap &
          SUMO parameter for empty space left when following a vehicle &
          From 1.0 to 4.0 meters \\
        sigma &
          SUMO parameter for driver imperfection with 0 denoting perfect driving, as per SUMO's default car-following model &
          From 0.1 to 1.0\\
        tau &
          SUMO parameter for modeling a driver's desired minimum time headway &
          From 0.1 to 3.0 seconds\\
        lcStrategic &
          SUMO parameter for eagerness for performing strategic lane changing, with 0 indicating no unnecessary lane-changing &
          From  0.1 to 3.0\\
        lcCooperative &
          SUMO parameter for willingness to perform cooperative lane changing, with lower values indicating reduced cooperation &
          From  0.1 to 1.0\\
        lcSpeedGain &
          SUMO parameter for eagerness to perform lane changing to gain speed, with higher values indicating more lane-changing &
          From 0.1 to 3.0\\
        speedFactor &
          SUMO parameter for controlling an individual's speeding behavior, as a multiplier applied to the speed limit. This allows individual vehicles to overspeed based on a normal distribution &
          Normal distribution with Means ranging from 1.0 to 1.5 and Standard Deviation from 0.1 to 2.0 \\

    \hline
    \end{tabulary}
    \end{adjustbox}
\caption{Dataset Generation Variability}
\label{table:datagen}
\end{table*}

\section{Experimental Results}
\label{expresults}
In this section, we evaluate the performance of our proposed fusion-based dynamic graph neural networks (FDGNN) for approximating travel time on arterial roads in urban traffic corridors. A total of 100,000 hours of simulation records is utilized to generate three datasets, each containing 50,000 samples. These datasets are constructed based on either real-world route files (\textit{Real-TMC dataset}) or a mixture of real and randomly generated route files (\textit{Mixed-TMC dataset}). Another experimental dataset (\textit{Real-TMC-short dataset}) is generated with shorter length of intervals over inflow waveforms to count traffic volumes at each phase of movement around intersections. The model is trained separately on each dataset and also on a mixture of both datasets to analyze its generalizability. 

To address the absence of a baseline, we train our proposed framework on various purposefully designed datasets and evaluate it under multiple intentional scenarios. The base model, FDGNN, is trained on the \textit{Real-TMC dataset}, which uses data aggregated into 15-minute input intervals, and its performance is compared against two baseline model variations. The first variant, termed FDGNN-Short, is trained on the \textit{Real-TMC-Short dataset}, with data aggregated into 5-minute input intervals. The second variant, referred to as FDGNN-Mixed, is trained on the \textit{Mixed-TMC dataset}, also using data aggregated into 15-minute input intervals.

The training dataset is split into 70\% for training, while the remaining 30\% is divided equally, with 15\% used for hyperparameter tuning and the other 15\% reserved for performance evaluation. Model performance is assessed by comparing the actual (fitted) and predicted normal PDFs, using widely accepted evaluation metrics:

\begin{itemize}
    




    \item \textbf{Normalized Root Mean Squared Error (NRMSE)}:
    \[
    \text{NRMSE} = \frac{\sqrt{\frac{1}{n} \sum_{i=1}^{n} \left( y_{\text{true},i} - y_{\text{pred},i} \right)^2}}{\max(y_{\text{true}}) - \min(y_{\text{true}})}
    \]

    \item \textbf{Hellinger Distance (HLD)}:
    \[
    \text{HLD} =  \frac{1}{\sqrt{2}} \sqrt{ \sum_{i=1}^{n} \left( \sqrt{y_{\text{true},i}} - \sqrt{y_{\text{pred},i}} \right)^2}
    \]

    \item \textbf{Standard Deviation Error (STD)}:
    \[
    \text{STD Error} = \left| \sigma_{\text{true}} - \sigma_{\text{pred}} \right|
    \]
    where \( \sigma_{\text{true}} \) and \( \sigma_{\text{pred}} \) are the standard deviations of the true and predicted distributions, respectively.
    
    \item \textbf{Mean Absolute Percentage Error (MAPE)}:
    \[
    \text{MAPE} = \frac{1}{n} \sum_{i=1}^{n} \left| \frac{y_{\text{true},i} - y_{\text{pred},i}}{y_{\text{true},i}} \right| \times 100
    \]
    
    where \(y_{\text{true},i}\) and \(y_{\text{pred},i}\) are 250 discrete sample values obtained from the probability density of the normal distribution of actual and predicted travel time respectively, computed over evenly spaced (i.e., 10 seconds) over the range of 0-2500 seconds of travel time through either eastbound or westbound direction.
\end{itemize}

Further, to thoroughly analyze the model’s performance across different traffic flow dynamics and control settings, we divide the dataset into smaller subsets based on predefined thresholds across three distinct scenarios. This allows for a more detailed examination of the model's behavior under varying conditions.
\subsection{Effect of Cycle Length}
We subset the test set based on cycle lengths. We bucket the cycle lengths into 3 buckets:
\begin{enumerate}
    \item Low: For cycle lengths smaller than 160 seconds
    \item Medium: For cycle lengths between 160 and less than 200 seconds
    \item High: For cycle lengths equal to and above 200 seconds
\end{enumerate}

\subsection{Effect of Traffic Volume}
We subset the test set based on corridor volumes i.e. the number of vehicles completing the corridor journey within the exemplar, along a certain direction. We bucket the volumes into 3 buckets:
\begin{enumerate}
    \item Low: For corridor volume smaller than 700 vehicles
    \item Medium: For corridor volume between 700 and less than 900 vehicles
    \item High: For corridor volume equal to and above 900 vehicles
\end{enumerate}

\subsection{Effect of Maximum Green Duration Percentage}
We subset the test set based on maximum green time percentage i.e. the percentage of maximum green time to the corridor-through phase, to the cycle length, along a certain direction. It represents what percent of the cycle could potentially be given to the corridor direction of flow. We bucket the ratio into 3 buckets:
\begin{enumerate}
    \item Low: For maximum green time percentage less than 25 \% 
    \item Medium: For maximum green time percentage between 25 \% and less than 50 \% 
    \item High: For maximum green time percentage equal to or more than 50 \% 
\end{enumerate}

\begin{figure*}[htbp]
\centering
\captionsetup{justification=raggedright,singlelinecheck=false}
\includegraphics[scale=0.7]{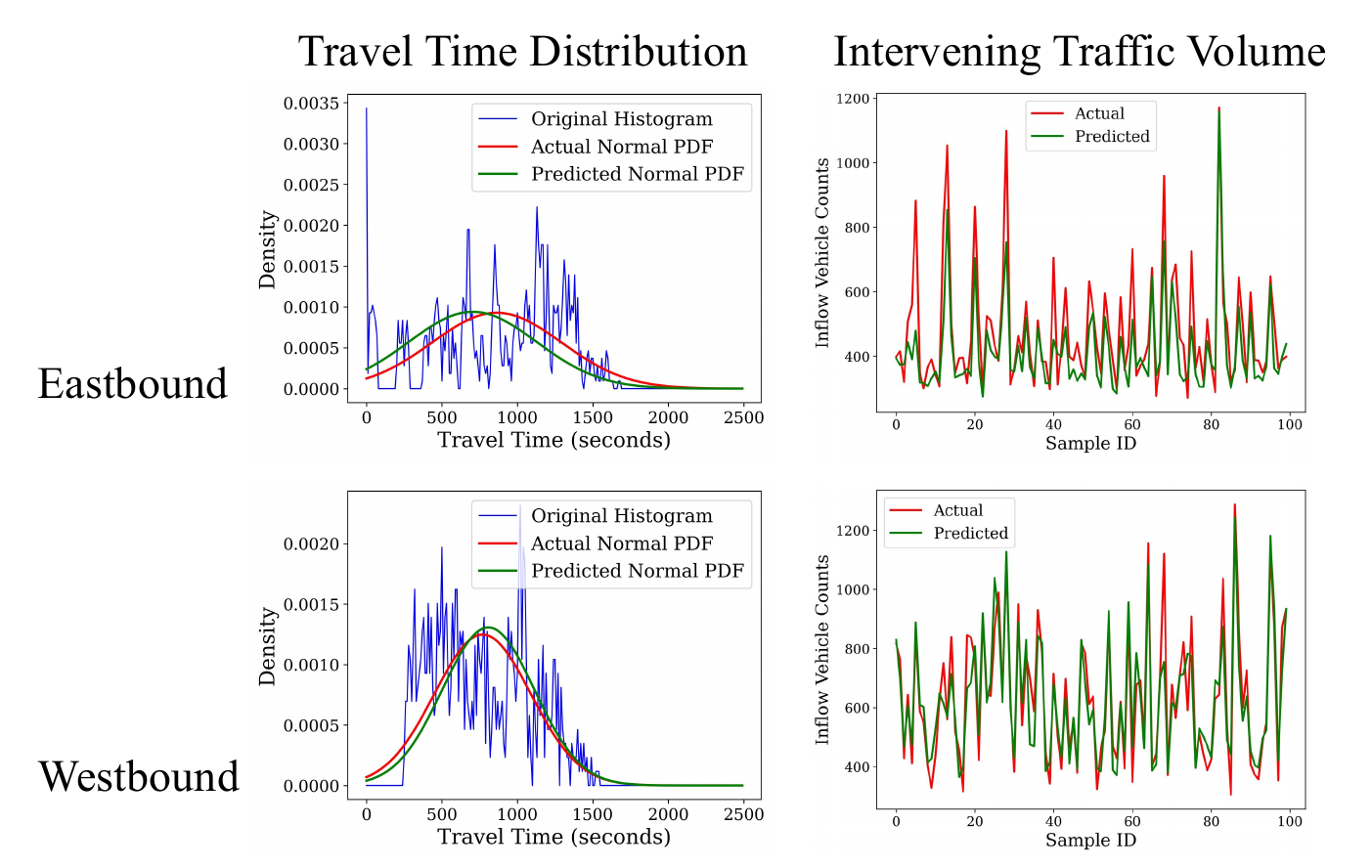}
\renewcommand{\figurename}{Figure}
\captionsetup{size=footnotesize}
\footnotesize
\caption{\textbf{Visualizing the results of FDGNN.} Comparison of actual (red) and predicted (green) curves by FDGNN for a single traffic scenario. The analysis includes the imputation of intervening inflow waveforms (right column) and the estimation of the normal Probability Density Function (PDF) of travel time in both eastbound and westbound arterial directions through the urban corridor.}

\label{fig:plots}
\end{figure*}

\begin{table*}[!htbp]
    \centering
    \resizebox{1\textwidth}{!}{%
    \begin{tabular}{@{}lccccccccccccc@{}}
        \toprule
        \multirow{2}{*}{.} & \multicolumn{1}{c}{} & \multicolumn{4}{c}{FDGNN} & \multicolumn{4}{c}{FDGNN-Short} & \multicolumn{4}{c}{FDGNN-Mixed} \\ 
        \cmidrule(lr){3-6} \cmidrule(lr){7-10} \cmidrule(lr){11-14}
        Experiment & Level & MAPE & STD & HLD & NRMSE & MAPE & STD & HLD & NRMSE & MAPE & STD & HLD & NRMSE \\ \midrule
        \multirow{3}{*}{Cycle Length} 
        & Low    & 0.02434770 & 17.03274536 & 0.09061160 & 0.07630504 & 0.02766613 & 23.52367783 & 0.10321835 & 0.06859140 & 0.02644867 & 21.89211082 & 0.10410079 & 0.06309731 \\ 
        & Medium & 0.01884929 & 24.90428543 & 0.07056899 & 0.05500218 & 0.02553535 & 38.01504898 & 0.09571076 & 0.06581541 & 0.02274527 & 25.90636635 & 0.08542714 & 0.04536204 \\ 
        & High   & 0.01401460 & 25.16983414 & 0.05289537 & 0.04202969 & 0.02309166 & 47.37397766 & 0.08680040 & 0.05483911 & 0.02090056 & 31.65452385 & 0.07882322 & 0.03523058 \\ 
        \midrule
        \multirow{3}{*}{Traffic Volume} 
        & Low    & 0.02166128 & 21.15789795 & 0.08088386 & 0.06759340 & 0.02727904 & 32.71726227 & 0.10197287 & 0.06273066 & 0.02979009 & 28.52194023 & 0.11562980 & 0.01453793 \\ 
        & Medium & 0.01945891 & 22.87958527 & 0.07274126 & 0.05478137 & 0.02557360 & 36.01268387 & 0.09573939 & 0.05737593 & 0.02021122 & 24.78332901 & 0.07633955 & 0.04835678 \\ 
        & High   & 0.01297529 & 20.92600632 & 0.04923087 & 0.03744519 & 0.02170557 & 37.73826599 & 0.08175873 & 0.05368090 & 0.01556541 & 27.27816772 & 0.05956285 & 0.03552472 \\ 
        \midrule
        \multirow{3}{*}{Maximum Duration \%} 
        & Low    & 0.01204259 & 26.28626442 & 0.04483104 & 0.04474065 & 0.02088253 & 44.97696686 & 0.07883339 & 0.07344692 & 0.02083357 & 30.60698891 & 0.07801799 & 0.04993816 \\ 
        & Medium & 0.01973771 & 22.00541878 & 0.07385662 & 0.05460556 & 0.02572506 & 34.94701767 & 0.09629201 & 0.05724831 & 0.02467635 & 26.59957314 & 0.09488192 & 0.01123968 \\ 
        & High   & 0.01850747 & 18.12760544 & 0.06880285 & 0.06923606 & 0.01954954 & 27.26508141 & 0.07472201 & 0.06617644 & 0.02273645 & 24.53370094 & 0.08540128 & 0.09216934 \\ 
        \midrule
        & Total  & 0.01972620 & 22.21240997 & 0.07379317 & 0.05436989 & 0.02586315 & 35.36487961 & 0.09672741 & 0.05630891 & 0.02456818 & 26.61972809 & 0.09455994 & 0.02681638 \\ 
        \bottomrule
    \end{tabular}%
    }
    \caption{\textbf{Performance evaluation.} FDGNN is examined with metrics such as MAPE, STD, HLD, and NRMSE when it is trained on real-world and/or randomly generated westbound traffic simulations. The analysis includes overall performance and scenario-specific divisions based on cycle length, traffic volume density, and maximum green duration ratio.}
    \label{table:error}
\end{table*}

The comparison provided in Table \ref{table:error} ensures a comprehensive assessment of the model's ability to generalize across diverse datasets and varying data aggregation resolutions. More specifically, it shows if its performance is robust on factual but also counterfactual scenarios, also if the performance is accurate for shorter periods of observation over inflow waveform time series. The upper (third) quartile, or 75th percentile, of travel time in the east and west directions, is computed as 900 seconds and 1200 seconds, respectively. Error measures are represented for various metrics used to evaluate the estimation of discrete probability density function (PDF) of the normal distribution for westbound travel time along arterial roads.

As indicated by the error values, the FDGNN model can approximate the distribution with a Standard Deviation Error (STD) of approximately 22.21 seconds. This means that, on average, the difference between the actual and predicted data from their mean values is only around 22 seconds. The Mean Absolute Percentage Error (MAPE) is approximately 0.02 times the upper quartile of the westbound travel time, which corresponds to a distance of around 24 seconds between the actual and predicted mean values. The Normalized Root Mean Squared Error (NRMSE) is about 0.05 times the range of the actual data (30–1200 seconds), suggesting that the model's estimation error is relatively very low. The Hellinger Distance (HLD) of 0.07 shows that the two probability density functions are nearly identical, though not exactly the same.

As observed, the performance of the FDGNN-Mixed model is very comparable to that of the original model, despite being trained on a mixture of real-world and randomly generated data. This indicates that the model can successfully handle not only factual but also counterfactual traffic scenarios, which is useful for decision-making and analysis by traffic engineers and policymakers.

When the window size for traffic volume counts in the waveform time series is reduced from 15 minutes to 5 minutes with cycle lengths ranging from 3-4 minutes, the results become less optimal still compared to the original model, as shown in the FDGNN-Mixed column. However, the deviation from the actual distribution's mean and standard deviation is around 35.36 seconds, and with an upper bound as large as 20 minutes, this deviation remains within an acceptable range.

The cycle length significantly impacts the model's performance, with a notable negative effect on both the mean and standard deviation of travel time estimates. Specifically, the model achieves better accuracy with shorter cycle lengths, as longer cycles introduce greater variability. This effect is more pronounced for FDGNN-Short, which relies on shorter input intervals that are less aligned with the traffic cycle length. As the cycle length increases from less than 160 seconds to over 200 seconds, FDGNN shows a deterioration of 12 seconds in mean estimation and 6 seconds in standard deviation estimation.

The impact of traffic volume on model performance depends on the input interval length. For FDGNN-Short, higher traffic volumes result in degraded performance, as the 5-minute input intervals fail to fully capture the dispersion of vehicle platoons. In contrast, FDGNN and FDGNN-Mixed, which use 15-minute input intervals, handle heavy traffic volumes more effectively, even when the traffic exceeds 900 vehicles per interval. This suggests that to achieve efficient and accurate real-time travel time estimation across varying traffic conditions throughout the day, adopting adaptable input intervals may be beneficial.

Additionally, the proportion of maximum green time negatively affects the outputs of all models, introducing greater variability in predictions for both the mean and standard deviation. This finding indicates that longer green times may disrupt the stability of traffic flow, leading to less accurate estimations of travel times. These observations highlight the need for carefully balanced signal timing plans and dynamic adjustment of model parameters to ensure reliable predictions across diverse traffic scenarios.

\section{Related Work}
\label{related}

Numerous computational techniques have been developed for optimizing signal timing along corridors, primarily aiming to achieve an optimal "progression" wave, enabling vehicles to travel as a synchronized platoon across multiple intersections. Examples include methods such as MAXBAND \cite{little1981maxband}, MULTIBAND \cite{stamatiadis1996multiband}, PASSER \cite{malakapalli1993enhancements}, TRANSYT \cite{wong2002group}, and SYNCHRO \cite{zou2004timing}. Additionally, \cite{christofa2016arterial} presents an integer programming-based algorithm that derives an optimal corridor timing plan based on pairwise intersection solutions.

Virtual Probe methods, introduced in \cite{Liu2009, Liu2012}, estimate the trajectory of an imaginary probe vehicle using kinematic equations as it traverses a signalized corridor. Loop detector data is employed to infer queue lengths at intersections, which influence the probe vehicle’s maneuvers. However, this approach only computes a single travel time for the corridor.

Building on this, \cite{liu2016iterative} integrates loop detector and probe vehicle data through an iterative Bayesian fusion method to estimate urban arterial travel times. Deep learning approaches have also been explored: \cite{nguyen2019deep} predicts travel speeds across road networks, while \cite{PETERSEN2019426} leverages trajectory data to forecast bus travel times. Alternatively, \cite{YANG2018325} proposes a modified Gaussian mixture model to estimate link travel time distributions along signalized arterials, combining vehicle re-identification, fixed-location magnetic sensors, and trajectory data.

However, despite these advancements, there remains limited research on estimating arterial travel time distributions using only loop detectors and signal timing data. This represents a significant challenge due to the inherent difficulty of vehicle re-identification within loop detector datasets.

\section{Conclusions and Future Work}
\label{conclusion}
In this paper, we delve into the challenge of approximating the distribution of arterial traveling time as a key metric for optimizing traffic corridor performance. 

We run the SUMO micro-simulator in a multi-threading manner to generate our extensive datasets. We extract the required information from structured XML log files and convert it to proper static and dynamic graph representation objects. The static and dynamic graph objects are designed to optimally and uniquely represent the evolving traffic dynamic state of urban traffic corridors. Each simulation generates structured XML log files processed into zip archives tailored for analysis. Metrics like vehicle counts, travel times, and density histograms are computed using Python modules. Floating Car Data (FCD) logs calculate eastbound and westbound arterial travel times by analyzing vehicle trajectories and timestamps to determine intersection entry and exit times.

The proposed Fusion-based Dynamic Graph Neural Networks (FDGNN) framework leverages interval-based traffic volumes and signal timing parameters to represent traffic corridors as dynamic graphs with time-evolving, direction-wise relationships. By employing an Attentional Graph Neural Network for graph completion and incorporating dynamic feature fusion, it learns global, context-aware representations, enhancing adaptability to complex traffic scenarios. The sequential learning framework fosters interdependent and hierarchical representation learning across modules, enabling more accurate and efficient traffic volume inference and improved performance on arterial corridors.

The performance of the model shows resilience and robustness while evaluated through a series of experiments. The experiments are designed to examine the effect of cycle length duration, traffic volume density, and the green time duration allocated to the estimation of bidirectional arterial travel time normal distributions. Furthermore, we demonstrate that our model can also successfully estimate under counterfactual traffic scenarios on randomly generated route files. 

This compact, deep learning-based approach, with a size of 230KB, is designed for urban corridors with any intersection topology, relying on only a few easily accessible traffic factors. Its scalability and robustness make it a promising solution for real-time traffic optimization and adaptive control, enhancing smart city infrastructure with efficient, corridor-level traffic management.

\section{Acknowledgments}
The work was supported in part by NSF CNS 1922782. The opinions, findings, and conclusions expressed in this publication are those of the authors and not necessarily those of NSF. The authors also acknowledge the University of Florida Research Computing for providing computational resources and support that have contributed to the research results reported in this publication.


\bibliographystyle{abbrv}

\bibliography{ref}
\newpage
\section{Biography Section}
\vskip -2\baselineskip plus -1fil
\begin{IEEEbiography}[{\includegraphics[width=1in,height=1.25in,clip,keepaspectratio]{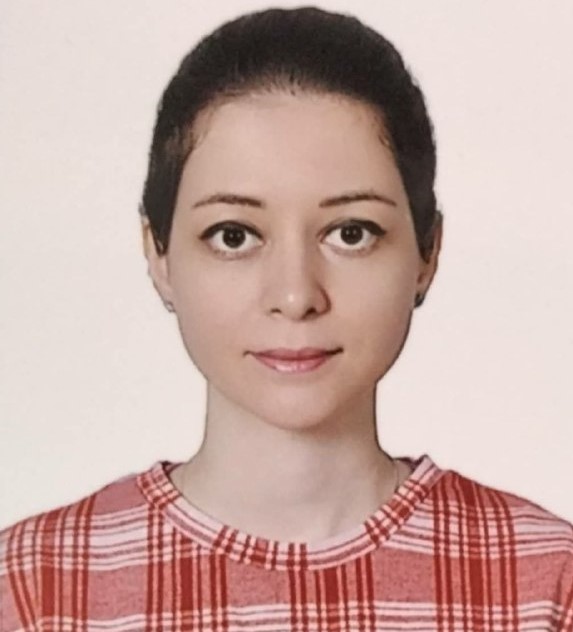}}]{Nooshin Yousefzadeh}
is currently pursuing her Ph.D. Degree with the Department of Computer and Information Science and Engineering, University of Florida, Gainesville, FL, USA. Her current research interests are in Explainable Artificial Intelligence and data-driven Machine Learning Algorithms for practical applications in Intelligent Transportation, Health Care, and Sustainability Science. 
\end{IEEEbiography}

\vskip -2\baselineskip plus -1fil
\begin{IEEEbiography}
[{\includegraphics[width=1in,height=1.25in,clip,keepaspectratio]{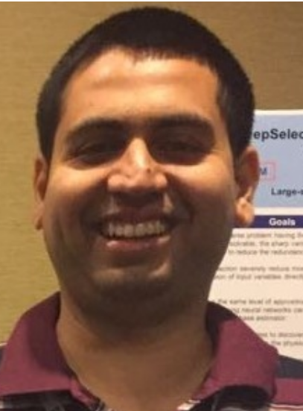}}]{Rahul Sengupta}
is a Ph.D. student at the Computer and Information Science Department at University of Florida, Gainesville, USA. His research interests include applying Machine Learning models to sequential and time series data, especially in the field of transportation engineering.
\end{IEEEbiography}

\vskip -2\baselineskip plus -1fil
\begin{IEEEbiography}[{\includegraphics[width=1in,height=1.25in,clip,keepaspectratio]{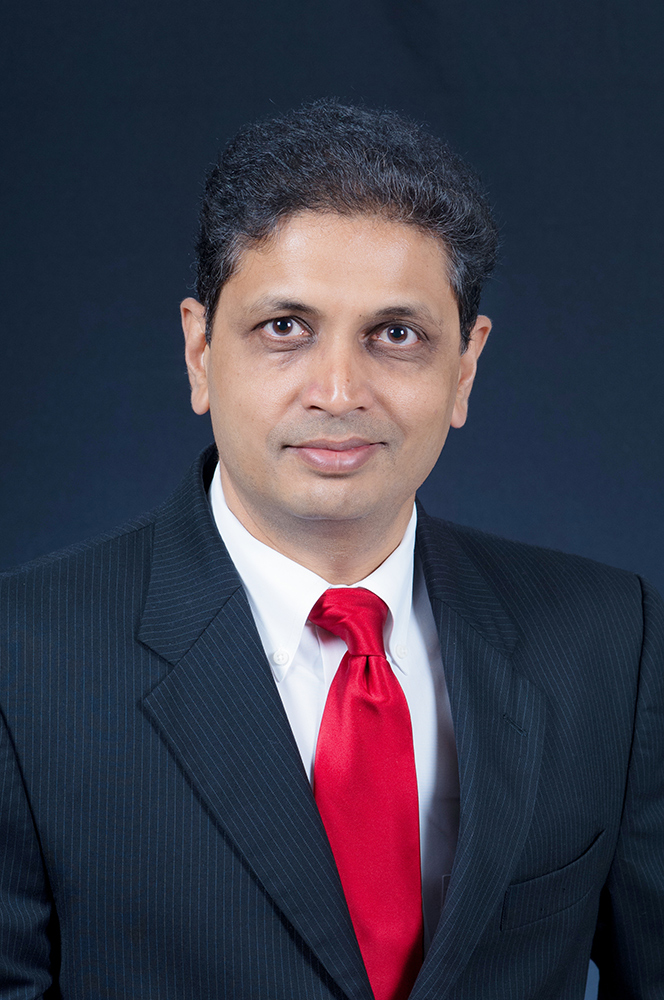}}]{Sanjay Ranka}
(Fellow, IEEE) is a Distinguished Professor in the Department of Computer Information Science and Engineering at University of Florida. His current research is on developing algorithms and software using Machine Learning, the Internet of Things, GPU Computing, and Cloud Computing for solving applications in Transportation and Health Care. He is a fellow of the IEEE, AAAS, and AIAA (Asia-Pacific Artificial Intelligence Association) and a past member of IFIP Committee on System Modeling and Optimization. He was awarded the 2020 Research Impact Award from IEEE Technical Committee on Cloud Computing. His research is currently funded by NIH, NSF, USDOT, DOE, and FDOT. From 1999-2002, as the Chief Technology Officer and co-founder of Paramark (Sunnyvale, CA), he conceptualized and developed a machine learning-based real-time optimization service called PILOT for optimizing marketing and advertising campaigns. Paramark was recognized by VentureWire/Technologic Partners as a Top 100 Internet technology company in 2001 and 2002 and was acquired in 2002.

\end{IEEEbiography}

\end{document}